\documentclass[conference]{IEEEtran}
\usepackage{stmaryrd}
\usepackage{amsfonts}

\usepackage{graphicx,times,amsmath} 

\IEEEoverridecommandlockouts    

\usepackage{color,calc,xcolor}
\usepackage{amssymb}

\newcommand{\bs}[1]{\mathbf{#1}}

\newcommand{\R}{\mathbb{R}}
\newcommand{\N}{\mathbb{N}}
\newcommand{\E}{\mathbf{E}}
\newcommand{\borel}{\mathcal{B}}
\newcommand{\bplus}{\borel^{+}}
\newcommand{\1}{\mathbf{1}}
\newcommand{\mleb}{\mu_{Leb}}
\newcommand{\X}{\bs{X}}
\newcommand{\Xt}[1][t]{\X_{#1}}
\newcommand{\Xtt}{\X_{t+1}}
\newcommand{\W}{\mathcal{W}}
\newcommand{\Wt}[1][t]{\mathcal{W}_{#1}}

\newcommand{\x}{\bs{x}}
\newcommand{\y}{\bs{y}}
\newcommand{\Y}{\bs{Y}}
\newcommand{\Yti}[1][i]{\Y_t^{#1}}
\newcommand{\Ytilde}[1][i]{\bs{\tilde{Y}}_t^{#1}}

\newcommand{\Wti}[1][i]{\bs{W}_t^{#1}}

\newcommand{\G}{\mathcal{G}}
\newcommand{\Gtil}{\tilde{\G}}
\newcommand{\pt}[1][t]{\bs{p}_{#1}}
\newcommand{\st}[1][t]{\sigma_{#1}}
\newcommand{\stt}{\sigma_{t+1}}

\newcommand{\Nti}[1][i]{\bs{N}_t^{#1}}
\newcommand{\Ntilde}[1][i]{\bs{\tilde{N}}_t^{#1}}
\newcommand{\Ntstar}[1][t]{\bs{\tilde{N}}_{#1}^\star}

\newcommand{\dt}[1][t]{\delta_{#1}}
\newcommand{\dmarkov}{\seq{\delta}}
\newcommand{\dpmarkov}{(\dt, \pt)_{t \in \N}}
\newcommand{\n}{\mathbf{n}}
\newcommand{\north}{\mathbf{n^{\perp}}}
\newcommand{\eone}{\mathbf{e}_{1}}
\newcommand{\etwo}{\mathbf{e}_{2}}
\newcommand{\ei}{\mathbf{e}_{i}}
\newcommand{\Nl}{\mathcal{N}}
\newcommand{\Nlun}{\mathcal{N}(0,1)}
\newcommand{\Nln}[1][n]{\mathcal{N}(\bs{0},\mathrm{Id}_{#1})}
\newcommand{\Nlambda}[1][\lambda]{\mathcal{N}_{#1:\lambda}}
\newcommand{\Ud}{\mathcal{U}}
\newcommand{\seq}[1]{\left( #1_t \right)_{t \in \N}}
\newcommand{\fdim}[1]{[#1]_1}

\newcommand{\cdfnn}[1][\delta]{\tilde{F}_{#1}}
\newcommand{\cdfnninv}[1][\delta]{\tilde{F}_{#1}^{-1}}
\newcommand{\ptilde}[1][\delta]{p_{#1}}
\newcommand{\ptildeone}[1][\delta]{p_{1,#1}}

\newcommand{\pstar}[1][\delta]{p_{#1}^\star}
\newcommand{\pstarone}[1][\delta]{p_{1,#1}^\star}
\newcommand{\pstartwo}[1][\delta]{p_{2,#1}^\star}
\newcommand{\cdfnone}[1][\delta]{F_{1,\delta}}
\newcommand{\ddomain}{\R_+}
\newcommand{\doet}[1][t]{\dt[#1]}

\newcommand{\doedo}{\ddomain}
\newcommand{\doe}{\delta}
\newcommand{\equald}{\overset{d}{=}}

\DeclareMathOperator{\argmax}{argmax}

\newtheorem{proposition}{Proposition}
\newtheorem{lemma}{Lemma}
\newtheorem{theorem}{Theorem}
\newtheorem{corollary}{Corollary}

\newenvironment{myproof}{\begin{proof}}{ \end{proof}} 

\begin{document}

\title{\ \\ \LARGE\bf Markov Chain Analysis of Evolution Strategies on a Linear Constraint Optimization Problem \thanks{Alexandre Chotard, Anne Auger and Nikolaus Hansen work in TAO, at INRIA-Saclay and LRI in University Paris-Sud, France (mail at name@lri.fr).} \thanks{Our thanks to Dirk Arnold for suggesting this work during PPSN is Sicily.}}

\author{Alexandre Chotard, Anne Auger and Nikolaus Hansen}


\maketitle

\begin{abstract}
This paper analyses a $(1,\lambda)$-Evolution Strategy, a randomised comparison-based adaptive search algorithm, on a simple constraint optimization problem. The algorithm uses resampling to handle the constraint and optimizes a linear function with a linear constraint. Two cases are investigated: first the case where the step-size is constant, and second the case where the step-size is adapted using path length control. 
We exhibit for each case a Markov chain whose stability analysis would allow us to deduce the divergence of the algorithm depending on its internal parameters. 
We show divergence at a constant rate when the step-size is constant. We sketch that with step-size adaptation geometric divergence takes place.
Our results complement previous studies where stability was assumed.
\end{abstract}


\section{Introduction}

Derivative Free Optimization (DFO) methods are  tailored for the optimization of numerical problems in a black-box context, where the algorithms can only query the objective function to optimize $f:\R^n \to \R$, and no properties on $f$, such as convexity or differentiability, is exploited. 

Evolution Strategies (ES) are comparison-based randomised DFO algorithms. At iteration $t$, solutions are sampled from a multivariate normal distribution centered in a vector $\Xt$. The candidate solutions are ranked according to $f$, and update of $\Xt$ and other parameters of the distribution (usually a step-size $\st$ and a covariance matrix) is performed using the ranking information given by the candidate solutions. Since ES do not directly use the function values of the new points, but only how $f$ ranks the different samples, they are invariant to the composition of the objective function by a strictly increasing function $h: \R\to\R$.

This property and the black-box scenario make Evolution Strategies suited for a wide class of real-world problems, where constraints on the variables are often given. Different techniques for handling constraints in randomised algorithms have been proposed, see \cite{CoelloCoello:2008} for a survey. For ES, common techniques are resampling, i.e. resample a solution till it lies in the feasible domain, repair of solutions that project unfeasible points onto the feasible domain (e.g. \cite{arnold2011repair}), penalty methods where unfeasible solutions are penalised either by a quantity that depends on the distance to the constraint (e.g. \cite{hansen2009tec} with adaptive penalty weights) (if this latter one can be computed) or by the constraint value itself (e.g. stochastic ranking \cite{runarsson2000stochastic}) or methods inspired from multi-objective optimization (e.g. \cite{mezura2005simple}).

In this paper we focus on the resampling method and study it on a simple constraint problem. More precisely, we study a $(1,\lambda)$-ES optimizing a linear function with a linear constraint and resampling any unfeasible solution until a feasible solution is sampled. The linear function models the situation where the current point is, relatively to the step-size, far from the optimum and ``solving'' this function means diverging. The linear constraint models being close to the constraint relatively to the step-size and far from other constraints. Due to the invariance of the algorithm to the composition of the objective function by a strictly increasing map, the linear function could be composed by a function without derivative and with many discontinuities without any impact on our analysis.

The problem we address was studied previously  for different step-size adaptation mechanisms: with constant step-size, self-adaptation and cumulative step-size adaptation \cite{arnold2011behaviour,arnold2012behaviour}. The drawn conclusion is that when adapting the step-size the $(1,\lambda)$-ES fails to diverge unless some requirements on internal parameters of the algorithm are met. However, the approach followed in the aforementioned studies relies on finding simplified theoretical models to explain the behaviour of the algorithm: typically those models arise by doing some approximations (considering some random variables equal to their expected value, ...) and assuming some mathematical properties like the existence of stationary distributions of underlying Markov chains. 

In contrast, our motivation is to study the real--in the sense not simplified--algorithm and prove rigorously different mathematical properties of the algorithm allowing to deduce the exact behaviour of the algorithm, as well as to provide tools and methodology for such studies. Our theoretical studies need to be complemented by simulations of the convergence/divergence rates. The mathematical properties that we derive show that these numerical simulations converge fast.

As for the step-size adaptation mechanism, our aim is to study the cumulative step-size adaptation (CSA), default step-size  mechanism for the CMA-ES algorithm \cite{cmaes}. The mathematical object to study for this purpose is a discrete time, continuous state space Markov chain that is defined as the couple:  evolution path and normalized distance to the constraint. More precisely stability properties like irreducibility, existence of a stationary distribution of this Markov chain need to be studied to deduce the geometric divergence of the CSA and have a rigorous mathematical framework to perform Monte Carlo simulations allowing to study the influence of different parameters of the algorithm. We start however by illustrating in details the methodology on the simpler case where the step-size is constant. We deduce in this case the divergence at a constant speed. We keep--due to some space limitation--the details of the generalization to the CSA study for a future publication and give only a sketch of the results.

This paper is organized as follows. In Section~\ref{sc:pr} we define the $(1,\lambda)$-ES using resampling and the problem. In Section~\ref{sec:preliminary} we provide some preliminary derivations on the distributions that come into play for the analysis. In Section~\ref{sc:cst} we analyze the constant step-size case: exhibit the Markov chain, prove its stability and deduce the divergence of the $(1,\lambda)$-ES on the constraint problem. In Section~\ref{sc:csa} we sketch out our results when the step-size is adapted using cumulative step-size adaptation. Finally we discuss our results and our methodology in Section~\ref{sc:discuss}.

\subsection*{Notations}

Throughout this article, we denote by $\varphi$ the density function of the standard multivariate normal distribution, and $\Phi$ the cumulative distribution function of a standard univariate normal distribution. The standard (unidimensional) normal distribution is denoted $\Nlun$, the ($n$-dimensional) multivariate  normal distribution with covariance matrix identity is denoted $\Nln$ and the $i^{\mathrm{th}}$ order statistic of $\lambda$ i.i.d. standard normal random variables is denoted $\Nlambda[i]$.  The uniform distribution on an interval $I$ is denoted $\Ud_I$. We denote $\mu_{Leb}$ the Lebesgue measure. The set of natural numbers (including $0$) is denoted $\N$, and the set of real numbers $\R$. We denote $\R_+$ the set $\lbrace x \in \R | x \geq 0 \rbrace$, and for $A \subset \R^n$, the set $A^*$ denotes $A \backslash \lbrace\bs{0}\rbrace$  and $\1_{A}$ denotes the indicator function of $A$. For two vectors $\x \in \R^n$ and $\bs{y} \in \R^n$, we denote $[\x]_i$ the $i^{\mathrm{th}}$-coordinate of $\x$, and $\x .\y$ the scalar product of $\x$ and $\y$. Take $(a,b) \in \N^2$ with $a\geq b$, we denote $[a..b]$ the interval of integers between $a$ and $b$. For a topological set $\mathcal{X}$, $\mathcal{B}(\mathcal{X})$ denotes the Borel algebra of $\mathcal{X}$. For $\X$ and $\Y$ two random vectors, we denote $\X \equald \Y$ if $\X$ and $\Y$ are equal in distribution. For $(X_t)_{t\in \N}$ a sequence of random variables and $X$ a random variable we denote $X_t \overset{a.s.}{\rightarrow} X$ if $X_t$ converges almost surely to $X$ and $X_t \overset{P}{\rightarrow} X$ if $X_t$ converges in probability to $X$.

\section{Problem statement and algorithm definition} \label{sc:pr}

\subsection{$(1,\lambda)$-ES with resampling}

In this paper, we study the behaviour of a $(1,\lambda)$-Evolution Strategy \emph{maximizing} a function $f$: $\R^n \rightarrow \R$, $\lambda \geq 2$, $n \geq 2$, with a constraint defined by a function $g : \R^n \rightarrow \R$ restricting the feasible space to $X_{\textrm{feasible}} = \lbrace \bs{x}\in \R^n | g(\bs{x}) \geq 0 \rbrace$. To handle the constraint, the algorithm resamples any unfeasible solution until a feasible solution is found.

From iteration $t \in \N$, given the vector $\Xt \in \R^{n}$ and step-size $\st \in \R_{+}^*$, the algorithm generates $\lambda$ new candidates:
\begin{equation}
\Yti = \Xt +\st \Nti \enspace,
\end{equation}
with $i \in [1 .. \lambda]$, and $(\Nti)_{i \in [1 .. \lambda]}$ i.i.d. standard multivariate normal random vectors. If a new sample $\Yti$ lies outside the feasible domain, that is $g(\Yti) < 0$, then it is resampled until it lies within the feasible domain. The first feasible $i^{\rm th}$ candidate solution is denoted $\Ytilde$ and the realization of the multivariate normal distribution giving $\Ytilde$ is $\Ntilde$, which is called a feasible step. Note that $\Ntilde$ is not distributed as a multivariate normal distribution, further details on its distribution are given later on.

We define $\star \! = \underset{i \in [1 .. \lambda]}{\argmax} f( \Ytilde )$ as the index realizing the maximum objective function, and call $\Ntstar$ the selected step. The vector $\Xt$ is then updated as the solution realizing the maximum value of the objective function, i.e.
\begin{equation} \label{eq:Xtt}
\Xtt = \Ytilde[\star] = \Xt + \st \Ntstar \enspace.
\end{equation}

The step-size and other internal parameters are then adapted. We denote for the moment in a non specific manner the adaptation as
\begin{equation} \label{eq:sig}
\stt = \st \xi_t
\end{equation}
where $\xi_t$ is a random variable whose distribution is a function of the selected steps $(\Ntstar[i])_{i \leq t}$. We will define later on specific rules for this adaptation.

\subsection{Linear fitness function with linear constraint}

\begin{figure}
\begin{center}
\includegraphics[width=0.43\textwidth]{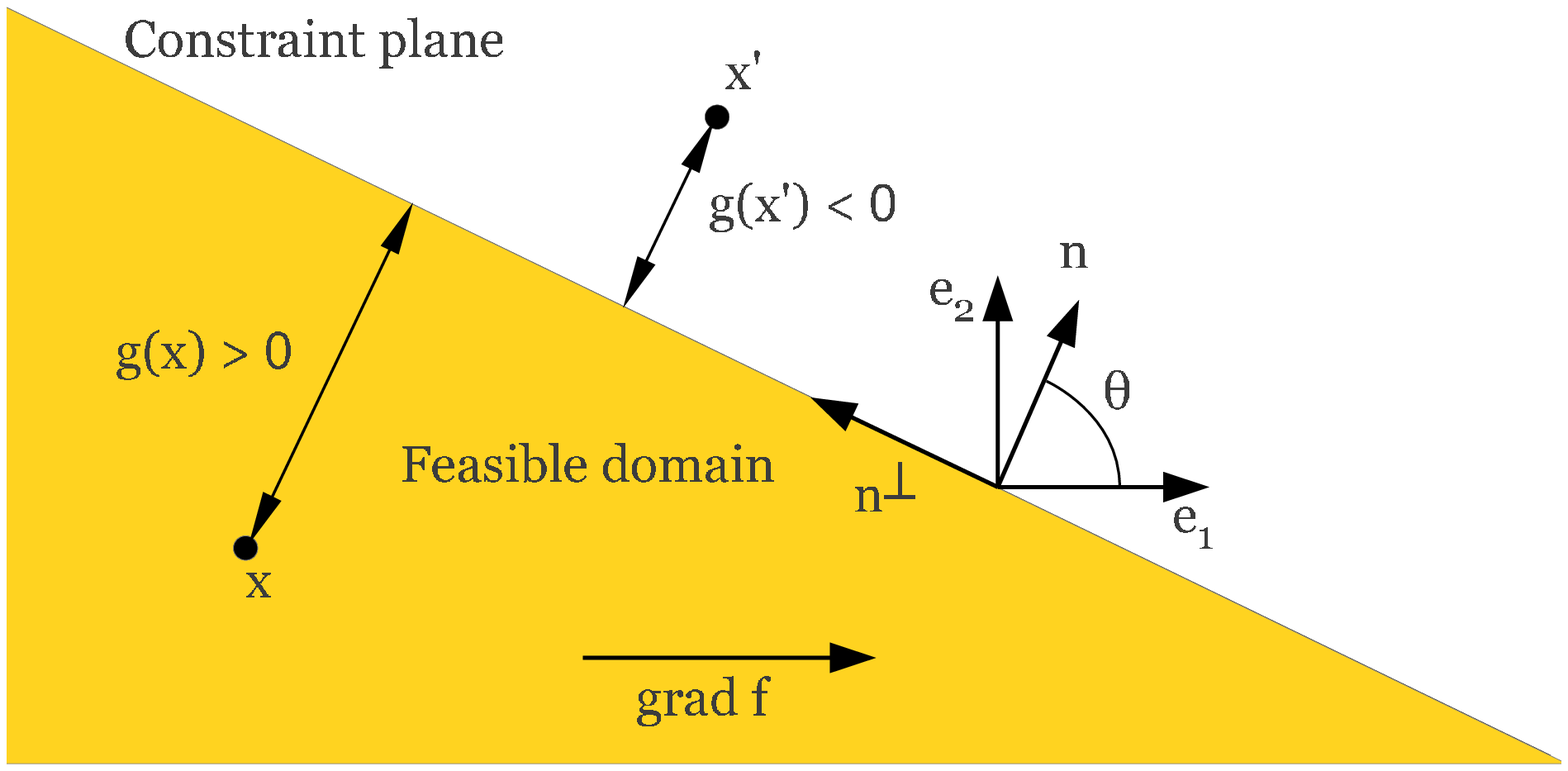}
\end{center}
\caption{Linear function with a linear constraint, in the plane generated by $\nabla f$ and $\n$, a normal vector to the constraint hyperplane with angle $\theta \in (0, \pi/2)$ with $\nabla f$. The point $\x$ is at distance $g(\x)$ from the constraint.}
\label{fg:problem}
\end{figure}

In this paper, we consider the case where $f$, the function that we optimize, and $g$, the constraint, are linear functions. W.l.o.g., we assume that $\| \nabla f\| = \| \nabla g \| = 1$. We denote $\n := - \nabla g $ a vector normal to the constraint hyperplane. We choose an orthonormal Euclidean coordinate system with basis $(\ei)_{i \in [1 ..n]}$ with its origin located on the constraint hyperplane where $\eone$ is equal to the gradient $\nabla f$, hence 
\begin{equation}\label{eq:f}
f(\x) = [\x]_{1}
\end{equation}
and the vector $\etwo$ lives in the plane generated by $\nabla f$ and $\n$ and is such that the angle between $\etwo$ and $\n$ is positive. We define $\theta$ the angle between $\nabla f$ and $\n$, and restrict our study to $\theta \in (0, \pi/2)$. The function $g$ can be seen as a signed distance to the linear constraint as
\begin{equation} \label{eq:g}
g(\x) = \x.\nabla g = -\x.\n  =  -[\x]_1 \cos \theta -[\x]_2 \sin \theta \enspace .
\end{equation}
A point is feasible if and only if $g(\x) \geq 0$ (see Figure~\ref{fg:problem}). Overall the problem reads
\begin{equation}\label{eq:pbdef}
\begin{split}
{\rm maximize}\,\,\,\, f(\x) = [\x]_{1}\,\,\,\, {\rm subject~to~}  \\ g(\x) = -[\x]_1 \cos \theta -[\x]_2 \sin \theta \geq 0 \enspace.
\end{split} 
\end{equation}

Although $\Ntilde$ and $\Ntstar$ are in $\R^n$, due to the choice of the coordinate system and  the independence of the sequence $([\Nti]_k)_{k \in [1..n]}$, only the two first coordinates of these vectors are affected by the resampling implied by $g$ and the selection according to $f$. Therefore $[\Ntstar]_k \sim \Nlun$ for $k \in [3..n]$. With an abuse of notations, the vector $\Ntilde$ will  denote the 2-dimensional vector $([\Ntilde]_1,[\Ntilde]_2)$, likewise $\Ntstar$ will also denote the 2-dimensional vector $([\Ntstar]_{1},[\Ntstar]_{2})$, and $\n$ will denote the 2-dimensional vector $(\cos \theta, \sin \theta)$. The coordinate system will also be used as $(\eone, \etwo)$  only.

Following \cite{arnold2011behaviour,arnold2012behaviour,arnold2008behaviour}, we denote the normalized signed distance to the constraint as $\dt$, that is 
\begin{equation}\label{eq:dt}
\dt = \frac{g(\Xt)}{\st} \enspace .
\end{equation}

We initialize the algorithm by choosing $\Xt[0] = -\n$ and $\st[0] = 1$, which implies that $\dt[0] = 1$.

\section{Preliminary results and definitions}\label{sec:preliminary}

Throughout this section we derive the probability density functions of the random vectors $\Ntilde$ and $\Ntstar$ and give a definition of $\Ntilde$ and of $\Ntstar$ as a function of $\dt$ and of an i.i.d. sequence of random vectors.

\subsection{Feasible steps} \label{sc:fea}

The random vector $\Ntilde$, the $i^{\rm th}$ feasible step, is distributed as the standard multivariate normal distribution truncated by the constraint, as stated in the following lemma.
\begin{lemma} \label{lm:fea}
Let a $(1,\lambda)$-ES with resampling optimize a function $f$ under a constraint function $g$. If $g$ is a linear form determined by a vector $\n$ as in \eqref{eq:g}, then the distribution of the feasible step $\Ntilde$ only depends on the normalized distance to the constraint $\dt$ and its density given that $\dt$ equals $\delta$ reads
\begin{equation} \label{eq:p}
\ptilde \left( \x  \right) = \frac{ \varphi(\x) \1_{\ddomain}\left( \delta - \x.\n \right)}{\Phi(\delta)} \enspace .
\end{equation}
\end{lemma}

\begin{myproof}
A solution $\Yti$ is feasible if and only if $g(\Yti) \geq 0$, which is equivalent to $-(\Xt+\st \Nti).\n \geq 0 $. Hence dividing by $\st$, a solution is feasible if and only if $\dt  = - \Xt.\n/\st \geq \Nti.\n $.
Since a standard multivariate normal distribution is rotational invariant, $\Nti.\n$ follows a standard (unidimensional) normal distribution. Hence the probability that a solution $\Yti$ or a step $\Nti$  is feasible is given by 
\begin{equation*} 
\Pr( \Nlun \leq \dt ) = \Phi\left(\dt\right) \enspace.
\end{equation*}
Therefore the density probability function of the random variable $\Ntilde.\n$ for $\dt = \delta$ is $x \mapsto \varphi(x) \1_{\ddomain}(\delta - x)/\Phi(\delta)$. For any vector $\north$ orthogonal to $\n$ the random variable $\Ntilde.\north$ was not affected by the resampling and is therefore still distributed as a standard (unidimensional) normal distribution. With a change of variables using the fact that the standard multivariate normal distribution is rotational invariant we obtain the joint distribution of Eq.~\eqref{eq:p}. 
\end{myproof}

Then the marginal density function $\ptildeone$ of $[\Ntilde]_{1}$ can be computed by integrating Eq.~\eqref{eq:p} over $[\x]_{2}$ and reads
\begin{equation} \label{eq:p1}
\ptildeone\left(x \right) = \varphi \left(x\right) \frac{\Phi\left(\frac{\delta - x \cos \theta}{\sin \theta}\right)}{\Phi\left(\delta \right)} \enspace ,
\end{equation}
(see \cite[Eq.~4]{arnold2011behaviour} for details) and we denote $\cdfnone$ its cumulative distribution function.

It will be important in the sequel to be able to express the vector $\Ntilde$ as a function of $\dt$ and of a \emph{finite} number of random samples. Hence we give an alternative way to sample $\Ntilde$ rather than the resampling technique that involves an unbounded number of samples.

\begin{lemma} \label{lm:Ntilde}
Let a $(1,\lambda)$-ES with resampling optimize a function $f$ under a constraint function $g$, where $g$ is a linear form determined by a vector $\n$ as in \eqref{eq:g}.
Let the feasible step $\Ntilde$ be the random vector described in Lemma~\ref{lm:fea} and $\bs{Q}$ be the 2-dimensional rotation matrix of angle $\theta$. Then 
\begin{equation}\label{def:Ntilde}
\Ntilde ~\equald~ \cdfnninv[\dt](U_{t}^{i}) \n + \mathcal{N}_{t}^{i} \north = \bs{Q}^{-1} \begin{pmatrix} \cdfnn[\dt]^{-1}(U_{t}^{i}) \\ \mathcal{N}_{t}^{i} \end{pmatrix}
\end{equation}
where $\cdfnninv[\dt]$ denotes the generalized inverse of the cumulative distribution of $\Ntilde . \n$ \footnote{The generalized inverse of $\cdfnn$ is $\cdfnninv[\dt](y) := \inf_{x\in\R}\lbrace \cdfnn[\dt](x) \geq y \rbrace$.}, $U_{t}^{i} \sim \Ud_{[0,1]}$, $\mathcal{N}_{t}^{i} \sim \Nlun$ with $(U_{t}^{i})_{i \in [1..\lambda], t \in \N}$ i.i.d. and $(\mathcal{N}_{t}^{i})_{i \in [1..\lambda], t \in \N}$ i.i.d. random variables. 
\end{lemma}

\begin{myproof}%
We define a new coordinate system $(\n, \north)$ (see Figure~\ref{fg:problem}). It is the image of $(\eone, \etwo)$ by  $\bs{Q}$. In the new basis $(\n, \north)$, only the coordinate along $\n$ is affected by the resampling. Hence the random variable $\Ntilde.\n$  follows a truncated normal distribution with cumulative distribution function $\cdfnn[\dt]$ equal to $\min(1,\Phi(x)/\Phi(\dt))$, while the random variable $\Ntilde.\north$ follows an independent standard normal distribution, hence $\Ntilde \equald (\Ntilde.\n)\n + \mathcal{N}_{t}^{i} \n^{\perp}$. Using the fact that if a random variable has a cumulative distribution $F$, then for $F^{-1}$ the generalized inverse of $F$, $F^{-1}(U)$ with $U \sim \Ud_{[0,1]}$ has the same distribution as this random variable, we get that $\cdfnn[\dt]^{-1}(U_{t}^{i}) \equald \Ntilde.\n$, so we obtain Eq.~\eqref{def:Ntilde}.
\end{myproof}

We now extend our study to the selected step $\Ntstar$.

\subsection{Selected step}

The selected step $\Ntstar$ is chosen among the different feasible steps $(\Ntilde)_{i \in [1..\lambda]}$ to maximize the function $f$, and has the density described in the following lemma.
\begin{lemma} \label{lm:pstar}
Let a $(1,\lambda)$-ES with resampling optimize the problem \eqref{eq:pbdef}. Then the distribution of the selected step $\Ntstar$ only depends on the normalized distance to the constraint $\dt$ and its density given that $\dt$ equals $\delta$ reads 
\begin{align} \label{eq:pstar}
\pstar \! \left(\x \right) \! &= \! \lambda \ptilde\left(\x \right) \cdfnone([\x]_1 )^{\lambda -1} \enspace , \\
 &= \! \lambda \frac{\varphi(\x) \1_{\R_+}(\delta - \x.\n)}{\Phi(\delta)} \!\left( \! \int_{-\infty}^{[\x]_1} \!\!\!\!\!\! \varphi(u) \frac{\Phi(\frac{\delta - u\cos\theta}{\sin \theta})}{\Phi(\delta)} \mathrm{d}u \!\right)^{\lambda-1} \nonumber
\end{align}
where $\ptilde$ is the density of $\Ntilde$ given that $\dt = \delta$ given in Eq.~\eqref{eq:p} and $\cdfnone$ the cumulative distribution function of $[\Ntilde]_1$ whose density is given in Eq.~\eqref{eq:p1} and $\n$ the vector $(\cos\theta, \sin\theta)$.
\end{lemma}

\begin{myproof}
The function $f$ being linear, the rankings on $(\Ntilde)_{i\in [1..\lambda]}$ corresponds to the order statistic on $([\Ntilde]_1)_{i\in [1..\lambda]}$. If we look at the joint cumulative distribution $F_\delta^\star$ of $\Ntstar$
\begin{align*}
F_\delta^\star(x,y) &= \Pr \left( [\Ntstar]_1 \leq x, [\Ntstar]_2 \leq y\right) \\
& = \sum_{i=1}^\lambda \Pr \left( \!  \Ntilde[i] \leq \left( \!\!\begin{array}{c} x \\ y \end{array} \!\! \right), [\Ntilde[j]]_1 < [\Ntilde[i]]_1 \text{ for } j \neq i  \right)
\end{align*}
by summing disjoints events. 
The vectors $(\Ntilde)_{i \in [1..\lambda]}$ being independent and identically distributed 
\begin{align*}
F_\delta^\star(x,y) &= \lambda \Pr \left( \Ntilde[1] \leq \left( \! \begin{array}{c} x \\ y \end{array} \! \right), [\Ntilde[j]]_1 < [\Ntilde[1]]_1 \text{ for } j \neq 1 \right) \\
& = \lambda \int_{-\infty}^x\int_{-\infty}^y p_\delta (u,v) \prod_{j=2}^\lambda \Pr([\Ntilde[j]]_1 < u) \mathrm{d}v \mathrm{d}u \\
& = \lambda \int_{-\infty}^x\int_{-\infty}^y p_\delta (u,v) \cdfnone(u)^{\lambda-1} \mathrm{d}v \mathrm{d}u \enspace .
\end{align*}
Deriving $F_\delta^\star$ on $x$ and $y$ yields the density of $\Ntstar$ of Eq.~\eqref{eq:pstar}.
\end{myproof}

We may now obtain the marginal of $[\Ntstar]_1$ and $[\Ntstar]_2$. 
\begin{corollary}
 Let a $(1,\lambda)$-ES with resampling optimize the problem \eqref{eq:pbdef}. Then the marginal distribution of $[\Ntstar]_1$ only depends of $\dt$ and its density given that $\dt$ equals $\delta$ reads
\begin{align} \label{eq:pstar1}
\pstarone \left(x \right) &= \lambda \ptildeone ( x ) \cdfnone(x )^{\lambda -1} \enspace , \\
&= \lambda \varphi(x) \frac{\Phi\left(\frac{\delta - x\cos\theta}{\sin\theta}\right)}{\Phi(\delta)} \cdfnone(x )^{\lambda -1} \enspace , \nonumber 
\end{align}
and the same holds for $[\Ntstar]_2$ whose marginal density reads
\begin{align} \label{eq:pstar2}
\pstartwo \left(y \right) &= \lambda \frac{\varphi(y)}{\Phi(\delta)} \int_{-\infty}^{\frac{\delta - y \sin \theta}{\cos \theta}} \varphi(u) \cdfnone(u)^{\lambda-1} \mathrm{d}u \enspace.
\end{align}
\end{corollary}

\begin{myproof}
Integrating Eq.~\eqref{eq:pstar} directly yields Eq.~\eqref{eq:pstar1}.

The conditional density function of $[\Ntstar]_2$ is 
$$
\pstartwo (y | [\Ntstar]_1 = x) = \frac{\pstar( (x,y) )}{ \pstarone(x )} \enspace .
$$
As $\pstartwo(y ) = \int_{\R} \pstartwo (y | [\Ntstar]_1 = x) \pstarone (x) \mathrm{d}x$, using the previous equation with Eq.~\eqref{eq:pstar} gives that $\pstartwo(y ) = \int_{\R} \lambda \ptilde((x,y) ) \cdfnone (x)^{\lambda-1} \mathrm{d}x$, which with Eq.~\eqref{eq:p} gives 
\begin{equation*}
 \pstartwo (y ) = \lambda \frac{\varphi(y)}{\Phi(\delta)}\int_{\R} \! \varphi(x)\1_{\R_+} \! \left(\delta - \left( \!\! \begin{array}{c} x \\ y \end{array} \!\! \right) .\n \right) \cdfnone(x)^{\lambda-1} \mathrm{d}x.
\end{equation*}
The condition $\delta - x\cos\theta - y\sin\theta \geq 0$ is equivalent to $x \leq (\delta - y\sin\theta)/\cos\theta$, hence Eq.~\eqref{eq:pstar2} holds.
\end{myproof}

We will need in the next sections an expression of the random vector $\Ntstar$ as a function of $\dt$ and a random vector composed of a \emph{finite} number of i.i.d. random variables. To do so, using notations of Lemma~\ref{lm:Ntilde}, we define the function $\Gtil : \ddomain\times([0,1]\times\R) \rightarrow \R^2$ as
\begin{equation} \label{eq:Gtilde}
\Gtil(\delta, \bs{w}) = \bs{Q}^{-1} \left( \begin{array}{c} \cdfnn^{-1}\left( [\bs{w}]_1 \right) \\ {[\bs{w}]_2} \end{array} \right) \enspace .
\end{equation}
According to Lemma~\ref{lm:Ntilde}, given that $U \sim \Ud_{[0,1]}$ and $\Nl \sim \Nlun$, $( \cdfnn^{-1}(U) , \Nl )$ (resp. $\G(\delta,(U,\Nl))$) is distributed as the resampled step $\Ntilde$ in the coordinate system $(\n, \n^\perp)$ (resp. $(\bs{e}_1, \bs{e}_2)$).
 Finally, let $(\bs{w}_i)_{i \in [1..\lambda]} \in ([0,1]\times \R)^\lambda$ and let $\G : \R_+\times([0,1]\times\R)^{\lambda} \rightarrow \R^2$ be the function defined as
\begin{equation} \label{eq:G}
\G(\delta, (\bs{w}_i)_{i \in [1..\lambda]}) = \underset{\bs{N} \in \left\lbrace \Gtil(\delta, \bs{w}_i) | i \in [1..\lambda] \right\rbrace}{\argmax} f(\bs{N}) \enspace .
\end{equation}
As shown in the following proposition, given that $\Wti \sim (\Ud_{[0,1]}, \Nlun)$ and $\Wt = (\Wti)_{i \in [1 .. \lambda]}$, the function $\G(\delta, \Wt)$ is distributed as the selected step $\Ntstar$.

\begin{proposition} \label{pr:g}
Let a $(1,\lambda)$-ES with resampling optimize the problem defined in Eq.~\eqref{eq:pbdef}, and let $(\Wti)_{i \in [1 .. \lambda], t \in \N}$ be an i.i.d. sequence of random vectors with $\Wti \sim (\Ud_{[0,1]}, \Nlun)$, and $\Wt = (\Wti)_{i \in [1 .. \lambda]}$. Then
\begin{equation}\label{eq:NtstarG}
\Ntstar  ~\equald~  \G(\dt, \Wt) \enspace,
\end{equation}
where the function $\G$ is defined in Eq.~\eqref{eq:G}.
\end{proposition}

\begin{myproof} 
Since $f$ is a linear function $f(\Ytilde) = f(\Xt) + \st f(\Ntilde)$, so $f(\Ytilde) \leq f(\Ytilde[j])$ is equivalent to $f(\Ntilde) \leq f(\Ntilde[j])$. Hence $\star = \argmax_{i \in [1..\lambda]} f(\Ntilde)$ and therefore $\Ntstar = \argmax_{\bs{N} \in \lbrace \Ntilde | i \in [1..\lambda] \rbrace } f(\bs{N})$. From Lemma~\ref{lm:Ntilde} and Eq.~\eqref{eq:Gtilde}, $\Ntilde \equald  \Gtil(\dt, \Wti)$, so $\Ntstar \equald \argmax_{\bs{N} \in \lbrace \Gtil(\dt, \Wti) | i \in [1..\lambda] \rbrace } f(\bs{N})$, which from \eqref{eq:G} is $\G(\dt, \Wt)$.
\end{myproof}

\section{Constant step-size case} \label{sc:cst}

We illustrate in this section our methodology analysis on the simple case where the step-size is constantly equal to $\sigma$ and prove that then $(\Xt)_{t\in\N}$ diverges almost surely at constant speed (Theorem~\ref{th:cstdiv}). The analysis of the CSA will then be a generalisation of the results presented here, with a few more technical results to derive.

As suggested in \cite{arnold2011behaviour}, the sequence $\dmarkov$ plays a central role for the analysis, and we will show that it admits a stationary measure. We first prove that this sequence is an homogeneous Markov chain.
\begin{proposition}
Consider the $(1,\lambda)$-ES with resampling and with constant step-size $\sigma$ optimizing the constraint problem~\eqref{eq:pbdef}. 
Then the sequence $\dt=g(\Xt)/\sigma$ is an homogeneous Markov chain on $\ddomain$ and
\begin{equation} \label{eq:cstmarkov}
\dt[t+1] = \dt - \Ntstar . \n ~\equald~ \dt - \G(\dt, \Wt).\n \enspace, 
\end{equation}
where $\G$ is the function defined in~\eqref{eq:G} and $(\Wt)_{t\in\N} = (\Wti)_{i \in [1..\lambda],t\in\N}$ is an i.i.d. sequence with $\Wti \sim (\Ud_{[0,1]}, \Nlun)$ for all $(i,t) \in [1..\lambda]\times \N$.
\end{proposition}
\begin{myproof}
It follows from the definition of $\dt$ that $\dt[t+1] = \frac{g\left(\Xtt\right)}{\st[t+1]} = \frac{-\left(\Xt + \sigma\Ntstar\right).\n}{\sigma} = \dt - \Ntstar . \n$, and in Proposition~\ref{pr:g} we state that $\Ntstar ~\equald~ \G(\dt, \Wt)$. Since $\dt[t+1]$ has the same distribution as a time independent function of $\dt$ and of $\Wt$ where $(\Wt)_{t \in \N}$ are i.i.d., it is an homogeneous Markov chain.
\end{myproof}

The Markov Chain $\dmarkov$ comes into play for investigating the divergence of $f(\Xt)=[\Xt]_{1}$. Indeed, we can express $\frac{[\Xt - \Xt[0]]_1}{t}$ in the following manner:
\begin{align}
\frac{[\Xt - \Xt[0]]_1}{t} & = \frac{1}{t} \sum_{k=0}^{t-1} [\X_{k+1}]_{1} - [\X_{k}]_{1} \nonumber
 \\ \label{eq:xrec} &= 
\frac{\sigma}{t}\sum_{k=0}^{t-1} [\Ntstar[k]]_{1} ~\equald~ \frac{\sigma}{t}\sum_{k=0}^{t-1} [\G(\dt[k],\W_{k})]_1 \enspace .
\end{align}
The latter term suggests the use of a Law of Large Numbers (LLN) to prove the convergence of $\frac{[\Xt - \Xt[0]]_1}{t}$ which will in turn imply--if the limit is positive--the divergence of $f(\Xt)$ at a constant rate. Sufficient conditions on a Markov chain to be able to apply the LLN include the existence of an invariant probability measure $\pi$. The limit term is then expressed as an expectation over the stationary distribution. More precisely, assume the LLN can be applied, the following limit will hold
\begin{align}
\lim_{t \to \infty}\frac{[\Xt - \Xt[0]]_1}{t} & = \sigma \int_{\R^{+}} \E  \left([\G(\delta,\W) ]_1\right) \pi(d \delta) \label{eq:xlln} \\ & =  \lim_{t \to \infty} \E_{\dt[0] \sim \mu}\left([\Xtt]_{1} - [\Xt]_{1}\right) \enspace,
\end{align}
with $\mu$ any initial distribution.
The latter term corresponds to the limit of the progress rate (see \cite[Eq.~2]{arnold2011behaviour}).
The invariant measure $\pi$ is also underlying  the study carried out in \cite[Section~4]{arnold2011behaviour} where more precisely it is stated: {\it ``Assuming for now that the mutation strength $\sigma$ is held constant, when the algorithm is iterated, the distribution of $\delta$-values tends to a stationary limit distribution.''}. We will now provide a formal proof that indeed $\dmarkov$ admits a stationary limit distribution $\pi$, as well as prove some other useful properties that will allow us in the end to conclude to the divergence of $(f(\Xt))_{t \in \N}$.

\subsection{Study of the stability of $\dmarkov$}

We study in this section the stability of $\dmarkov$. We first derive its transition kernel $P(\delta,A) := \Pr( \dt[t+1] \in A | \dt[t] = \delta) $ for all $\delta \in \ddomain$ and $A \in \borel(\ddomain)$. Since
$\Pr( \dt[t+1] \in A | \dt = \delta)  = \Pr(\delta_{t} - \Ntstar.\n \in A | \dt = \delta ) \enspace,
$
\begin{equation}\label{trans:kernel}
P(\delta,A) = \int_{\R^2} \1_A \left( \delta - \bs{u}.\n \right) \pstar \left( \bs{u}\right) \mathrm{d}\bs{u}
\end{equation}
where $\pstar$ is the density of $\Ntstar$ given in \eqref{eq:pstar}. For $t \in \N^*$, the $t$-step transition  kernel $P^t$ is defined by $P^t(\delta, A) := \Pr(\dt \in A | \dt[0] = \delta)$.

From the transition kernel, we will now derive the first properties on the Markov chain $\dmarkov$. First of all we investigate the so-called $\psi$-irreducible property.

A Markov chain $(\doet)_{t\in\N}$ on a state space $\doedo$ is  \emph{$\psi$-irreducible} if there exists a non-trivial measure $\psi$ such that for all set $A \in \borel(\doedo)$ with $\psi(A)>0$ and for all $\doe \in \doedo$, there exists $t \in \N^*$ such that $P^t(\doe,A)>0$. We denote $\bplus(\doedo)$ the set of Borel sets of $\doedo$ with strictly positive $\psi$-measure.

We also need the notion of \emph{small sets}: a set $C \in \borel(\doedo)$ is called a small set if there exists $m \in \N^*$ and a non trivial measure $\nu_m$ such that for all set $A \in \borel(\doedo)$ and all $\delta \in C$
\begin{equation}
P^m(\doe,A) \geq \nu_m(A) \enspace .
\end{equation}
If there exists $C$ a $\nu_1$-small set such that $\nu_1(C) > 0$ then the Markov chain is said \emph{strongly aperiodic}.

\begin{proposition} \label{pr:cstirreducible}
Consider a $(1,\lambda)$-ES with resampling and with constant step-size optimizing the constraint problem \eqref{eq:pbdef}  and let $\dmarkov$ be the Markov chain exhibited in \eqref{eq:cstmarkov}. Then $\dmarkov$ is $\mleb$-irreducible, strongly aperiodic, and compact sets are small sets.
\end{proposition}

\begin{myproof}
Using Eq.~\eqref{trans:kernel} and Eq.~\eqref{eq:pstar} the transition kernel can be written
\begin{equation*}
P(\delta, A) \! = \! \lambda \! \int_{\R^2} \!\! \1_A (\delta - \left( \!\! \begin{array}{c} x \\ y \end{array} \!\! \right).\n ) \frac{\varphi(x)\varphi(y)}{\Phi(\delta)} \cdfnone( x )^{\lambda-1} \mathrm{d}y\mathrm{d}x \enspace .
\end{equation*}
We remove $\delta$ from the indicator function by a substitution of variables  $u = \delta - x\cos\theta -y\sin\theta$, and $v = x\sin\theta -y\cos\theta$. As this substitution is the composition of a rotation and a translation the determinant of its Jacobian matrix is $1$. We denote $h_\delta : (u,v) \mapsto (\delta-u)\cos\theta + v\sin\theta$, $h_\delta^\perp : (u,v) \mapsto (\delta-u)\sin\theta - v\cos\theta$ and $g (\delta,u,v) \mapsto \lambda \varphi(h_\delta(u,v))\varphi(h_\delta^\perp(u,v))/\Phi(\delta)\cdfnone(h_\delta(u,v))^{ \lambda -1}$. Then $x = h_\delta(u,v)$, $y = h_\delta^\perp(u,v)$ and
\begin{equation}
\label{eq:ker}
P(\delta, A) = \int_{\R}  \int_\R 1_A(u) g(\delta,u,v)  \mathrm{d}v \mathrm{d}u \! \enspace .
\end{equation}
For all $\delta,u,v$ the function $g(\delta,u,v)$ is strictly positive hence for all $A$ with $\mleb(A) > 0$, $P(\delta,A) > 0$. Hence $\dmarkov$ is irreducible with respect to the Lebesgue measure.

In addition, the function $(\delta,u,v) \mapsto g(\delta,u,v)$ is continuous as the composition of continuous functions (the continuity of $\delta \mapsto F_{1,\delta}(x)$ for all $x$ coming from the  dominated convergence theorem). Given a compact $C$ we hence know that there exists $g_C > 0$ such that for all $(\delta,u,v) \in C\times[0,1]^2$, $g(\delta,u,v) \geq g_C > 0$.
Hence for all $\delta \in C$,
$$
P(\delta,A) \geq \underbrace{g_C  \mleb(A \cap [0,1]) }_{:=\nu_C(A)} \enspace.
$$
The measure $\nu_C$ being non-trivial, the previous equation shows that compact sets are small and that for $C$ a compact such that $\mleb(C \cap [0,1]) > 0$, we have $\nu_C(C) > 0$ hence the chain is strongly aperiodic.
\end{myproof}

The application of the LLN for a $\psi$-irreducible Markov chain $(\doet)_{t \in \N}$ on a state space $\doedo$ requires the existence of an \emph{invariant measure} $\pi$, that is satisfying for all $A \in \borel(\doedo)$
\begin{equation} \label{eq:positive}
\pi(A) = \int_{\doedo} P(\doe, A) \pi(\mathrm{d}\doe)  \enspace .
\end{equation}
If a Markov chain admits an invariant probability measure then the Markov chain is called positive.

A typical assumption to apply the LLN is positivity and Harris-recurrence.
A $\psi$-irreducible chain $(\doet)_{t \in \N}$ on a state space $\doedo$ is \emph{Harris-recurrent} if for all set $A \in \bplus(\doedo)$ and for all $\doe \in \doedo$, $\Pr ( \eta_A = \infty | \doe_0 = \doe ) = 1$ where $\eta_A$ is the occupation time of A, i.e. $\eta_A = \sum_{t=1}^\infty 1_A(\doet)$.
We will show that the Markov chain $\dmarkov$ is positive and Harris-recurrent by using so-called Foster-Lyapunov drift conditions: define the \emph{drift} operator for a positive function $V$ as 
$$
\Delta V(\delta) = \E[ V(\dt[t+1]) | \dt=\delta ] - V(\delta)\enspace.
$$
Drift conditions translate that outside a small set, the drift operator is negative. We will show a drift condition for V-geometric ergodicity where given a function $f \geq 1$, a positive and Harris-recurrent chain $(\doet)_{t \in \N}$ with invariant measure $\pi$ is called \emph{$f$-geometrically ergodic} if $\pi(f) < \infty$ and there exists $r_f> 1$ such that
\begin{equation} \label{eq:geo}
\sum_{t \in \N} r_f^t \| P^t(\doe, \cdot) - \pi \|_f < \infty \enspace , \forall \doe  \in  \doedo \enspace ,
\end{equation}
where for $\nu$ a signed measure $\| \nu \|_f$ denotes $\sup_{g : |g| \leq f} | \int_{\doedo} g(x)\nu(\textrm{d}x) |$.

To prove $V$-geometric ergodicity, we will prove that there exists a small set $C$, constants $b \in \R$, $\epsilon \in \R_+^*$ and a function $V \geq 1$ finite for at least some $\doe_0 \in \doedo$ such that for all $\doe \in \doedo$
\begin{equation} \label{eq:V4}
\Delta V(\delta) \leq -\epsilon V(\delta) + b \1_{C}(\delta) \enspace .
\end{equation}
If the Markov chain $\dmarkov$ is $\psi$-irreducible and aperiodic, this drift condition implies that the chain is $V$-geometrically ergodic \cite[Theorem 15.0.1]{markovtheory}\footnote{The condition $\pi(V) < \infty$ is given by \cite[Theorem 14.0.1]{markovtheory}.}  as well as positive and Harris-recurrent\footnote{The function $V$ of \eqref{eq:V4} is unbounded off petite sets \cite[Lemma 15.2.2]{markovtheory}, hence with \cite[Theorem 9.1.8]{markovtheory} the Markov chain is Harris-recurrent.}.

Because compacts are small sets and drift conditions investigate the negativity outside a small set, we need to study the chain for $\delta$ large. The following lemma is a technical lemma studying the limit of $\E(\exp(\G(\delta, \W).\n))$  for $\delta$ to infinity.

\begin{lemma} \label{lm:farNtstar}
Consider the $(1,\lambda)$-ES with resampling optimizing the constraint problem \eqref{eq:pbdef}, and let $\G$ be the function defined in \eqref{eq:G}.
We denote $K$ and $\bar{K}$ the random variables $\exp(\G(\delta, \W).(a,b))$ and $\exp(a|[\G(\delta,\W)]_1| + b|[\G(\delta,\W)]_2|)$. For $\W \sim (\Ud_{[0,1]}, \Nl(0,1))^\lambda$ and any $(a,b)\in \R^2$ $\lim_{\delta \rightarrow +\infty} \E(K) = \E(\exp(a\Nlambda))\E(\exp(b \Nlun)) < \infty$ and $\lim_{\delta \rightarrow +\infty} \E(\bar{K})  < \infty$
\end{lemma}
For the proof see the appendix. We are now ready to prove a drift condition for geometric ergodicity.

\begin{proposition} \label{pr:csterg}
Consider a $(1,\lambda)$-ES with resampling and with constant step-size optimizing the constraint problem \eqref{eq:pbdef} and let $\dmarkov$ be the Markov chain exhibited in \eqref{eq:cstmarkov}. The Markov chain $\dmarkov$ is $V$-geometrically ergodic with $V : \delta \mapsto \exp(\alpha \delta)$ for $\alpha > 0$ small enough, and is Harris-recurrent and positive with invariant probability measure $\pi$.
\end{proposition}

\begin{myproof}
Take the function $V : \delta \mapsto \exp(\alpha\delta)$ then 
$\Delta V (\delta) = \E\left(\exp\left(\alpha\left(\delta - \G(\delta,\W).\n \right)\right)\right) - \exp\left(\alpha \delta \right) $,
$\frac{\Delta V}{V}(\delta)	= \E\left(\exp\left( - \alpha\G(\delta,\W).\n \right)\right) - 1 $.
With Lemma~\ref{lm:farNtstar} we obtain
$
\underset{\delta\rightarrow + \infty}{\lim} \E\left(\exp\left( - \alpha\G(\delta,\W).\n \right)\right) = \\ \E\left( \exp (-\alpha \Nlambda \cos\theta) \right) \E(\exp(-\alpha \Nlun \sin \theta)) < \infty \enspace .
$
As the right hand side of the previous equation is finite we can invert integral with series with Fubini's theorem, so with Taylor series the limit equals to
\begin{equation*}
\left( \! \sum_{i \in \N} \! \frac{\left(-\alpha \cos\theta\right)^i \E \! \left(\Nlambda ^i\right)}{i!} \!\! \right) \!\! \left( \! \sum_{i \in \N} \! \frac{\left(-\alpha \sin\theta \right)^i \E \! \left(\Nlun^i\right)}{i!} \!\! \right) \!\! \enspace ,
\end{equation*}
which in turns yields
\begin{align*}
\lim_{\delta \rightarrow + \infty} \!\! \frac{\Delta V}{V}(\delta) &= \! \left(1 - \alpha\E(\Nlambda)\cos\theta + o(\alpha)\right)\left(1 + \! o(\alpha)\right) \! - \! 1 \\
&= -\alpha\E(\Nlambda)\cos\theta + o(\alpha) \enspace .
\end{align*}
Since for $\lambda \geq 2$, $\E(\Nlambda) > 0$, for $\alpha >0$ and small enough we get $\lim_{\delta \rightarrow + \infty} \frac{\Delta V}{V}(\delta) < -\epsilon < 0$. Hence there exists $\epsilon > 0$, $M>0$ and $b\in \R$ such that
\begin{equation*}
\Delta V(\delta) \leq -\epsilon V(\delta) + b \1_{[0,M]}(\delta) \enspace .
\end{equation*}

According to Proposition~\ref{pr:cstirreducible}, $[0,M]$ is a small set, hence it is petite \cite[Proposition 5.5.3]{markovtheory}. Furthermore $\dmarkov$ is a $\psi$-irreducible aperiodic Markov chain so $\dmarkov$ satisfies the conditions of Theorem~15.0.1 from \cite{markovtheory}, which with Lemma~15.2.2, Theorem~9.1.8 and Theorem~14.0.1 of \cite{markovtheory} proves the proposition.
\end{myproof}

We now proved rigorously the existence (and unicity) of an invariant measure $\pi$ for the Markov chain $\dmarkov$, which provides the so-called steady state behaviour in \cite[Section 4]{arnold2011behaviour}. As the Markov chain $\dmarkov$ is positive and Harris-recurrent we may now apply a Law of Large Numbers \cite[Theorem 17.1.7]{markovtheory} in Eq~\eqref{eq:xrec} to obtain the divergence of $f(\Xt)$ and an exact expression of the divergence rate.

\begin{theorem} \label{th:cstdiv} 
Consider a $(1,\lambda)$-ES with resampling and with constant step-size optimizing the constraint problem \eqref{eq:pbdef} and let $\dmarkov$ be the Markov chain exhibited in \eqref{eq:cstmarkov}. 
The sequence  $(\fdim{\Xt})_{t \in \N}$ diverges in probability to $+\infty$ at constant speed, that is
\begin{align} \label{eq:cstdiv}
\frac{\fdim{\Xt - \Xt[0]}}{t} & \overset{P}{\underset{t \rightarrow +\infty}{\longrightarrow}} \sigma  \E_{\pi\times\mu_\W} \left( [\G\left(\delta, \W \right)]_1\right)  > 0 \enspace , 
\end{align}
with $\G$ defined in \eqref{eq:G} and $\W = (\bs{W}^i)_{i \in [1..\lambda]}$ where $(\bs{W}^i)_{i \in [1..\lambda]}$ is an i.i.d. sequence such that $\bs{W}^i \sim (\Ud_{[0,1]}, \Nlun)$ and $\mu_\W$ is the probability measure of $\W$.
\end{theorem}

\begin{myproof} From Proposition~\ref{pr:csterg} the Markov chain $\dmarkov$ is  Harris-recurrent and positive, and since $(\Wt)_{t \in \N}$ is i.i.d., the chain $(\dt, \Wt)$ is also Harris-recurrent and positive with invariant probability measure $\pi \times \mu_\W$, so to apply the Law of Large Numbers \cite[Theorem 17.0.1]{markovtheory} to $[\G]_1$ we only need $[\G]_1$ to be $\pi\times\mu_\W$-integrable.

With Fubini-Tonelli's theorem $\E_{\pi\times \mu_\W}(|[\G(\delta,\W )]_1|) $ equals to $ \E_{\pi} (\E_{\mu_\W}(|[\G(\delta,\W )]_1| ))$. As $\delta\geq 0$, we have $\Phi(\delta) \geq \Phi(0) = 1/2$, and for all $x\in\R$  as  $\Phi(x) \leq 1$, $\cdfnone(x) \leq 1$ and $\varphi(x) \leq \exp(-x^2/2)$ with Eq.~\eqref{eq:pstar1} we obtain that $|x| \pstarone(x) \leq 2 \lambda |x| \exp(-x^2/2)$ so the function $x \mapsto |x| \pstarone(x)$ is integrable. Hence for all $\delta \in \R_+$, $\E_{\mu_\W}(|[\G(\delta,\W )]_1| )$ is finite. Using the dominated convergence theorem, the function $\delta \mapsto \cdfnone(x)$ is continuous, hence so is $\delta \mapsto \pstarone(x)$. From \eqref{eq:pstar1} $|x|\pstarone(x) \leq 2 \lambda |x| \varphi(x)$,  which is integrable, so the dominated convergence theorem implies that the function $\delta \mapsto \E_{\mu_\W}(|[\G(\delta,\W]_1|)$ is continuous. Finally, using Lemma~\ref{lm:farNtstar} with Jensen's inequality shows that $\lim_{\delta \rightarrow +\infty} \E_{\mu_\W}(|[\G(\delta, \W)]_1|)$ is finite. Therefore the function  $\delta \mapsto \E_{\mu_\W}(|[\G(\delta,\W]_1|)$ is bounded by a constant $M \in \R_+$. As $\pi$ is a probability measure $\E_{\pi} (\E_{\mu_\W}(|[\G(\delta,\W )]_1| )) \leq M < \infty$, meaning $[\G]_1$ is $\pi\times\mu_\W$-integrable. Hence we may apply the LLN on Eq.~\eqref{eq:xrec}
\begin{equation*}
\frac{\sigma}{t} \sum_{k=0}^{t-1} [\G(\dt[k],\Wt[k])]_1 \overset{a.s.}{\underset{t \rightarrow + \infty}{\longrightarrow}} \sigma \E_{\pi\times\mu_\W}\left([\G(\delta,\W )]_1\right) < \infty \enspace .
\end{equation*}
The equality in distribution in \eqref{eq:xrec} allows us to deduce the convergence in probability of the left hand side of \eqref{eq:xrec} to the right hand side of the previous equation.

As the measure $\pi$ is an invariant measure for the Markov chain $\dmarkov$, using \eqref{eq:cstmarkov}, $\E_{\pi\times\mu_\W}(\delta) = \E_{\pi\times\mu_\W}(\delta-\G(\delta,\W).\n)$, hence $\E_{\pi\times\mu_\W}(\G(\delta,\W).\n) = 0$ and thus
\begin{equation*}
\E_{\pi\times\mu_\W}\left([ \G(\delta, \W)]_1 \right) = - \tan \theta \E_{\pi\times\mu_\W}\left([\G(\delta, \W)]_2\right) \enspace .
\end{equation*}
We see from Eq.~\eqref{eq:pstar2} that for $y > 0$, $\pstartwo(y ) < \pstartwo(-y)$  hence the expected value $\E_{\pi\times\mu_\W}([\G(\delta, \W)]_2)$ is strictly negative. With the previous equation it implies that $\E_{\pi\times\mu_\W}([\G(\delta, \W)]_1)$ is strictly positive.

\end{myproof}

We showed rigorously the divergence of $[\Xt]_1$ and gave an exact expression of the divergence rate, which is the limit of the progress rate defined in \cite[Eq.~(2)]{arnold2011behaviour}. The fact that the chain $\dmarkov$ is $V$-geometrically ergodic gives that $\sum_t r_V^t \|P^t(\delta,\cdot) - \pi\|_V < \infty$. This implies that the distribution $\pi$ can be simulated efficiently by a Monte Carlo simulation allowing to have precise estimations of the divergence rate of $[\Xt]_1$. Assuming a CLT could be applied, confidence intervals on the Monte Carlo simulations could also be obtained.

\begin{figure}\centering
\includegraphics[width=0.45\textwidth, trim=0 0ex 0 5ex 0, clip]{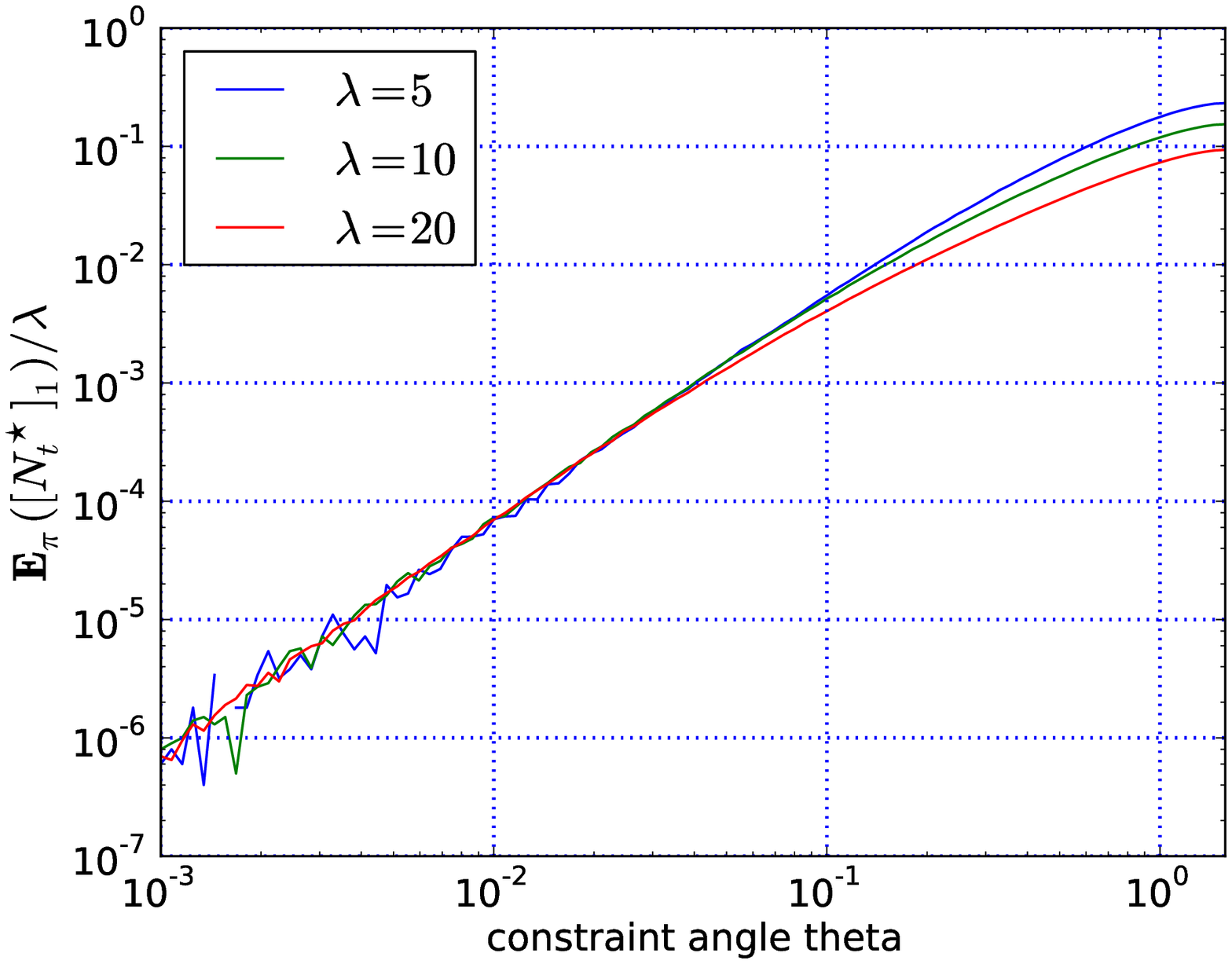}
\caption{Normalized progress rate $\varphi^\star = \E([\Ntstar]_1)$ divided by  $\lambda $ for the $(1,\lambda)$-ES with constant step-size and resampling, plotted against the constraint angle $\theta$, for $\lambda \in \lbrace 5, 10, 20 \rbrace$.}
\label{fg:cstpvst}
\end{figure}

A Monte Carlo simulation of the right hand side of Eq.~\eqref{eq:cstdiv} for $10^6$ time steps gives the progress rate $\varphi^\star =  \E([\Xtt - \Xt]_1)$, which once normalized by $\sigma$ and $\lambda$ yields Fig.~\ref{fg:cstpvst}. We normalize per $\lambda$ as in evolution strategies the cost of the algorithm is assumed to be the number of $f$-calls. We see that for small values of $\theta$, the normalized serial progress rate assumes roughly $\varphi^\star/\lambda \approx \theta^2$. Only for larger constraint angles the serial progress rate depends on $\lambda$ where smaller $\lambda$ are preferable.

\begin{figure}\centering
\includegraphics[width=0.45\textwidth, trim=0 0ex 0 5ex 0, clip]{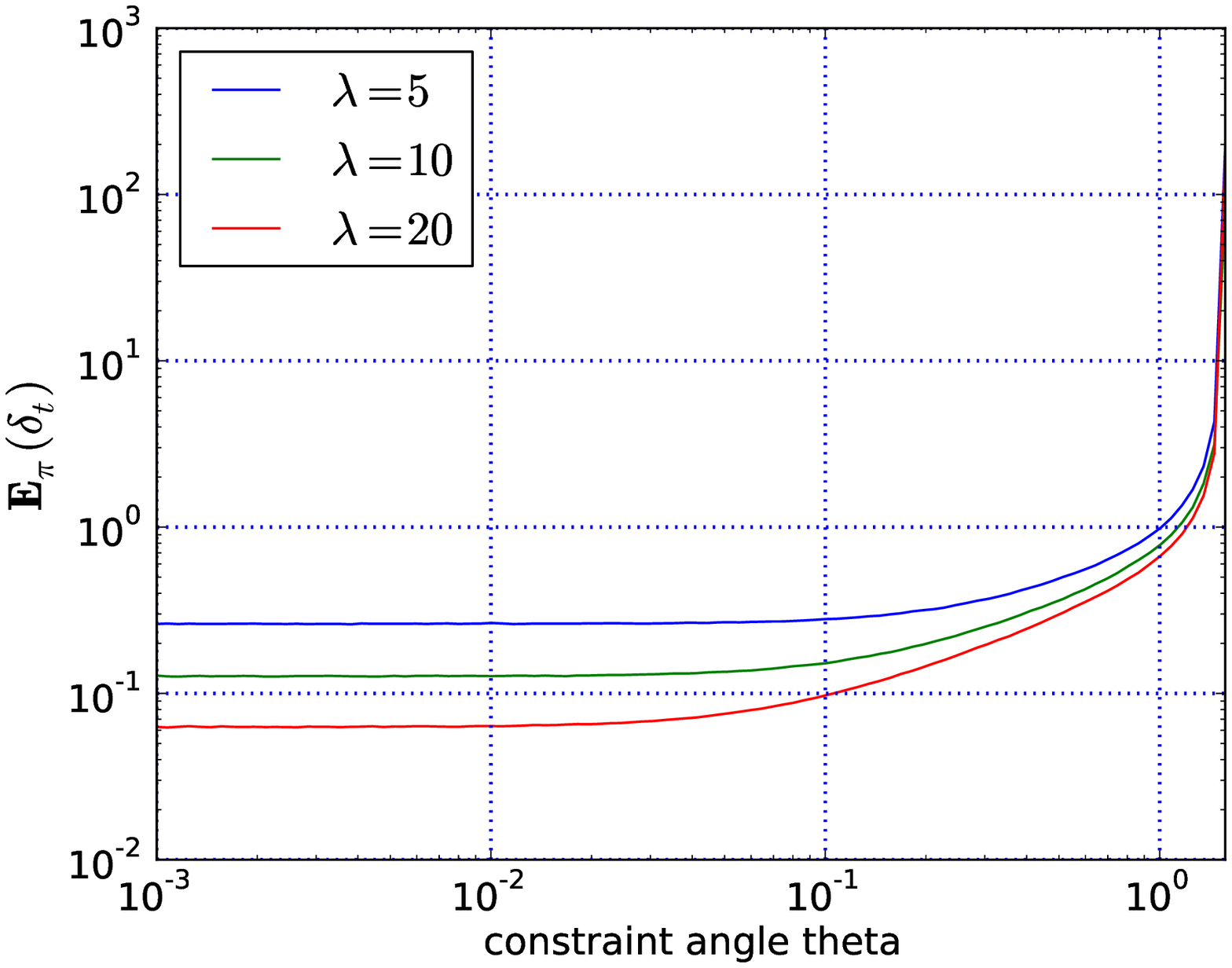}
\caption{Average normalized distance $\delta$ from the constraint for the $(1,\lambda)$-ES with constant step-size and resampling plotted against the constraint angle $\theta$ for $\lambda \in \lbrace 5, 10, 20 \rbrace$.}
\label{fg:cstdvst}
\end{figure}

Fig.~\ref{fg:cstdvst} is obtained through simulations of the Markov chain $\dmarkov$ defined in Eq.~\eqref{eq:cstmarkov} for $10^6$ time steps where the values of $\dmarkov$ are averaged over time. We see that when $\theta \rightarrow \pi/2$ then $\E_\pi ( \dt ) \rightarrow + \infty$ since the selection does not attract $\Xt$ towards the constraint anymore, while the resampling still repels $\Xt$ from the constraint. With a larger population size the algorithm is closer to the constraint, as better samples are more likely to be found close to the constraint.

\section{Cumulative Step-size Adaptation case} \label{sc:csa}
We generalise the previous results to the cumulative step-size adaptation mechanism. However due to space limitation we only sketch the results that we plan to present in details in an extended version of the paper. CSA introduces a new variable, $\pt$, called the evolution path. It is a weighted recombination of the previous selected steps, where the weight of $\Ntstar[k]$ is proportional to $(1-c)^{t-1-k}$ with $c \in (0,1]$ being the cumulation parameter. For $c=1$ the algorithm has "no memory" and the evolution path  $\pt$ is $\Ntstar[t-1]$. The step-size is adapted depending on the norm of $\pt$ \cite{cmaes}. The Markov chain to study in this case is $\dpmarkov$, except when $c=1$ where it is $\dmarkov$.

As in Section~\ref{sc:cst} if the Markov chain is $\psi$-irreducible, aperiodic, and compact sets are small, then for $c=1$ the Markov chain $\dmarkov$ is positive, Harris recurrent and $V$-geometrically ergodic, and a LLN can be applied on $\ln(\st/\st[0])$ to obtain that 
\begin{equation} \label{eq:csa}
\frac{1}{t} \ln \left( \frac{\st}{\st[0]} \right) \overset{a.s.}{\underset{t \rightarrow \infty}{\longrightarrow}} \frac{\left(  \E_{\pi_c\times\mu_\W} \left(\|\G\left(\delta, \W\right)\|^2\right) - 2 \right)}{2d_{\sigma} n} \enspace ,
\end{equation}
with $\pi_c$ the stationary measure of $\dmarkov$, $\G$ defined in \eqref{eq:G}, $\W = (\bs{W}^i)_{i \in [1..\lambda]}$ where $(\bs{W}^i)_{i \in [1..\lambda]}$ is an i.i.d. sequence such that $\bs{W}^i \sim (\Ud_{[0,1]}, \Nlun)$ and $\mu_\W$ the probability measure of $\W$. So the step-size converges (resp. diverges) exponentially fast when the right hand side of Eq.~\eqref{eq:csa} is strictly negative (resp. strictly positive).

\section{Discussion}\label{sc:discuss}

We investigated the $(1,\lambda)$-ES with constant step-size optimizing a linear function under a linear constraint handled by resampling unfeasible solutions. We prove the stability (formally V-geometric ergodicity) of the Markov chain $\dmarkov$ defined as the normalised distance to the constraint, which was \emph{pressumed} in \cite{arnold2011behaviour}. This property implies the divergence of the algorithm at a constant speed (see Theorem~\ref{th:cstdiv}). In addition, it ensures (fast) convergence of Monte Carlo simulations of the divergence rate, justifying their use.

We believe that with the same approach, the CSA can be analysed. Simulations suggest that geometric divergence occurs for a small enough cumulation parameter, $c$, or large enough population size, $\lambda$. However, smaller values of the constraint angle seem to increase the difficulty of the problem arbitrarily, i.e.\ no given values for $c$ and $\lambda$ solve the problem for \emph{every} $\theta \in (0,\pi/2)$.

Using a different covariance matrix to generate new samples can be interpreted as a change of the constraint angle. Therefore a correct adaptation of the covariance matrix will render the problem arbitrarily close to the one with $\theta=\pi/2$. The unconstrained linear function case has been shown to be solved by a $(1,\lambda)$-ES with cumulative step-size adaptation for a population size larger than $3$, regardless of other internal parameters \cite{chotard2012TRcumulative}. 
We believe this is a strong argument for using covariance matrix adaptation with ES when dealing with constraints, as pure step-size adaptation has been shown to be liable to fail on even a very basic problem.

This work provides a methodology that can be applied to many ES variants. It demonstrates that a rigorous analysis of the constrained problem can be achieved.  It relies on the theory of Markov chains for a continuous state space that once again proves to be a natural theoretical tool for analysing ESs, complementing particularly well previous studies \cite{arnold2011behaviour,arnold2012behaviour,arnold2008behaviour}.

\section*{Acknowledgments}This work was supported by the grants  ANR-2010-COSI-002 (SIMINOLE) and ANR-2012-MONU-0009 (NumBBO) of the French National Research Agency.



%

\def\V{\rm vol.~}
\def\N{no.~}
\def\pp{pp.~}
\def\Pot{\it Proc. }
\def\IJCNN{\it International Joint Conference on Neural Networks\rm }
\def\ACC{\it American Control Conference\rm }
\def\SMC{\it IEEE Trans. Systems\rm , \it Man\rm , and \it Cybernetics\rm }

\def\handb{ \it Handbook of Intelligent Control: Neural\rm , \it
    Fuzzy\rm , \it and Adaptive Approaches \rm }

\bibliographystyle{plain}
\bibliography{biblio}

\section*{Appendix} \label{sc:appendix}

Proof of Lemma~\ref{lm:farNtstar}.
\begin{myproof}
From Proposition~\ref{pr:g} the density probability function of $\G(\delta, \W)$ is $\pstar$, and from Eq.~\eqref{eq:pstar}
\begin{equation*}
 \pstar\!\left( \!\! \left( \!\!\! \begin{array}{c} x \\ y \end{array} \!\!\! \right) \!\! \right) = \lambda\frac{\varphi(x)\varphi(y)\1_{\R_+} \! \left(\delta - \left( \!\!\! \begin{array}{c} x \\ y \end{array} \!\!\! \right).\n\right)}{\Phi(\delta)} \cdfnone(x)^{\lambda - 1} \enspace .
\end{equation*}
From Eq.~\eqref{eq:p1} $\ptildeone(x) = \varphi(x) \Phi((\delta - x \cos\theta)/\sin\theta)/\Phi(\delta)$, so as $\delta \geq 0$ we have $1 \geq \Phi(\delta) \geq \Phi(0) = 1/2$, hence $\ptildeone(x) \leq 2\varphi(x)$. So $\ptildeone(x)$ converges when $\delta \rightarrow +\infty$ to $\varphi(x)$ while being bounded by $2\varphi(x)$ which is integrable. Therefore we can apply Lebesgue's dominated convergence theorem: $\cdfnone$ converges to $\Phi$ when $\delta \rightarrow + \infty$ and is finite.

For $\delta \in \ddomain$ and $(x,y)\in\R^2$ let $h_{\delta,y}(x)$ be $\exp(a x) \pstar((x,y))$. With Fubini-Tonelli's theorem  $\E(\exp( \G(\delta, \W).(a,b))) = \int_\R \int_\R \exp(b y) h_{\delta,y}(x) \mathrm{d}x \mathrm{d}y$. For $\delta \rightarrow +\infty$, $h_{\delta,y}(x)$ converges to $\exp(a x) \lambda \varphi(x) \varphi(y) \Phi(x)^{\lambda-1}$ while being dominated by $2\lambda \exp (a x)\varphi(x)\varphi(y)$, which is integrable. Therefore by the dominated convergence theorem and as the density of $\Nlambda$ is $x \mapsto \lambda \varphi(x)\Phi(x)^{\lambda-1}$, when $\delta \rightarrow +\infty$, $\int_{\R}h_{\delta,y}(x)\mathrm{d}x$ converges to $ \varphi(y) \E(\exp(a \Nlambda)) < \infty$.

So the function $y \mapsto \exp(b y) \int_{\R}h_{\delta,y}(x)\mathrm{d}x$ converges to $y \mapsto \exp(b y) \varphi(y) \E(\exp (a \Nlambda ))$ while being dominated by $y \mapsto 2\lambda\varphi(y)\exp(by) \int_\R \exp(ax)\varphi(x)\mathrm{d}x$ which is integrable. Therefore we may apply the dominated convergence theorem: $\E( \exp( \G(\delta, \W).(a,b) ) )$ converges to $\int_{\R} \exp(b y) \varphi(y) \E(\exp(a \Nlambda) ) \mathrm{d}y$ which equals to $\E(\exp(a \Nlambda))\E(\exp(b\Nlun))$; and this quantity is finite.

The same reasoning gives 
that $\lim_{\delta \to \infty}\E(\bar{K}) < \infty$.
\end{myproof}


\end{document}